\def\year{2022}\relax
\documentclass[letterpaper]{article} % DO NOT CHANGE THIS
\usepackage{aaai22}  % DO NOT CHANGE THIS
\usepackage{times}  % DO NOT CHANGE THIS
\usepackage{helvet}  % DO NOT CHANGE THIS
\usepackage{courier}  % DO NOT CHANGE THIS
\usepackage[hyphens]{url}  % DO NOT CHANGE THIS
\usepackage{graphicx} % DO NOT CHANGE THIS
\usepackage{natbib}  % DO NOT CHANGE THIS AND DO NOT ADD ANY OPTIONS TO IT
\usepackage{caption} % DO NOT CHANGE THIS AND DO NOT ADD ANY OPTIONS TO IT
\DeclareCaptionStyle{ruled}{labelfont=normalfont,labelsep=colon,strut=off} % DO NOT CHANGE THIS
\usepackage{algorithm}
\usepackage{algorithmic}

\usepackage{newfloat}
\usepackage{listings}
\floatstyle{ruled}
\newfloat{listing}{tb}{lst}{}
\floatname{listing}{Listing}
\usepackage{times}
\usepackage{latexsym}
\usepackage{graphicx}
\usepackage{array}
\usepackage{float}
\usepackage{enumitem}
\usepackage{amsfonts}
\usepackage{amsmath}
\usepackage[T1]{fontenc}
\usepackage{xcolor}

\usepackage[utf8]{inputenc}

\usepackage{microtype}

\usepackage[normalem]{ulem}

\newcommand{\zhihua}[1]{#1}

\newcommand{\pmark}{($p<0.05$)}
\title{NumGPT: Improving Numeracy Ability of Generative Pre-trained Models}
\author {
    Zhihua Jin,\textsuperscript{\rm 1}
    Xin Jiang, \textsuperscript{\rm 2}
    Xingbo Wang, \textsuperscript{\rm 1}
    Qun Liu,\textsuperscript{\rm 2}
    Yong Wang, \textsuperscript{\rm 3}
    Xiaozhe Ren, \textsuperscript{\rm 2}
    Huamin Qu \textsuperscript{\rm 1}
}
\affiliations {
    \textsuperscript{\rm 1} Department of Computer Science \& Engineering, Hong Kong University of Science and Technology, Hong Kong\\
    \textsuperscript{\rm 2} Huawei Noah's Ark Lab\\
    \textsuperscript{\rm 3} School of Computing and Information Systems, Singapore Management University, Singapore\\
    \{zjinak,xwangeg,huamin\}@ust.hk, \{Jiang.Xin,qun.liu,renxiaozhe\}@huawei.com,  yongwang@smu.edu.sg
}
\usepackage{bibentry}
\begin{document}

\maketitle

\begin{abstract}
Existing generative pre-trained language models (e.g., GPT) focus on modeling the language structure and semantics of general texts. However, those models do not consider the numerical properties of numbers and cannot perform robustly on numerical reasoning tasks (e.g., math word problems and measurement estimation). In this paper, we propose NumGPT, a generative pre-trained model that explicitly models the numerical properties of numbers in texts. Specifically, it leverages a prototype-based numeral embedding to encode the \textit{mantissa} of the number and an individual embedding to encode the \textit{exponent} of the number. A numeral-aware loss function is designed to integrate numerals into the pre-training objective of NumGPT. We conduct extensive experiments on four different datasets to evaluate the numeracy ability of NumGPT. The experiment results show that NumGPT outperforms baseline models (e.g., GPT and GPT with DICE) on a range of numerical reasoning tasks such as measurement estimation, number comparison, math word problems, and magnitude classification. Ablation studies are also conducted to evaluate the impact of pre-training and model hyperparameters on the performance.
\end{abstract}

\section{Introduction}
Pre-trained models such as GPT~\cite{radford2018improving,radford2019language,brown2020language} and BERT~\cite{devlin2018bert,liu2019roberta,lan2019albert} have made remarkable achievements in natural language processing. Through fully utilizing a large-scale unlabeled corpus, they have successfully gained state-of-the-art results in various kinds of natural language understanding tasks, such as GLUE~\cite{wang2018glue} and SuperGLUE~\cite{wang2019superglue}. 
% Moreover, with the development of computational resources, it has become possible to train deep neural network models with a large parameter size such as GPT-3~\cite{brown2020language} and Switch transformer~\cite{fedus2021switch}. These models have achieved significantly better performances in different NLP tasks, including news generation, question answering and so on.
%\yong{Zhihua, pls further check it.}

% larger models can be trained, and GPT-3~\cite{brown2020language} or Switch transformer~\cite{fedus2021switch} impress the researchers in NLP and even in the AI field. 

% 1. GPT-3 is very great LMs. However, it has poor numeracy performance when encountering large numberes. 
%Large scale pretrained language models also have obtain inspiring results in other tasks. 
%GPT-3~\cite{brown2020language} has successful ability to model language and obtains inspiring results. 
%GPT-3~\cite{brown2020language} also demonstrates that it is also a few shot learners for addition task. However, a limitation is observed. As the numbers increase in the question, the performance of GPT-3 drops down seriously. 

\begin{figure}[htb]
\centering 
\includegraphics[width=0.48\textwidth]{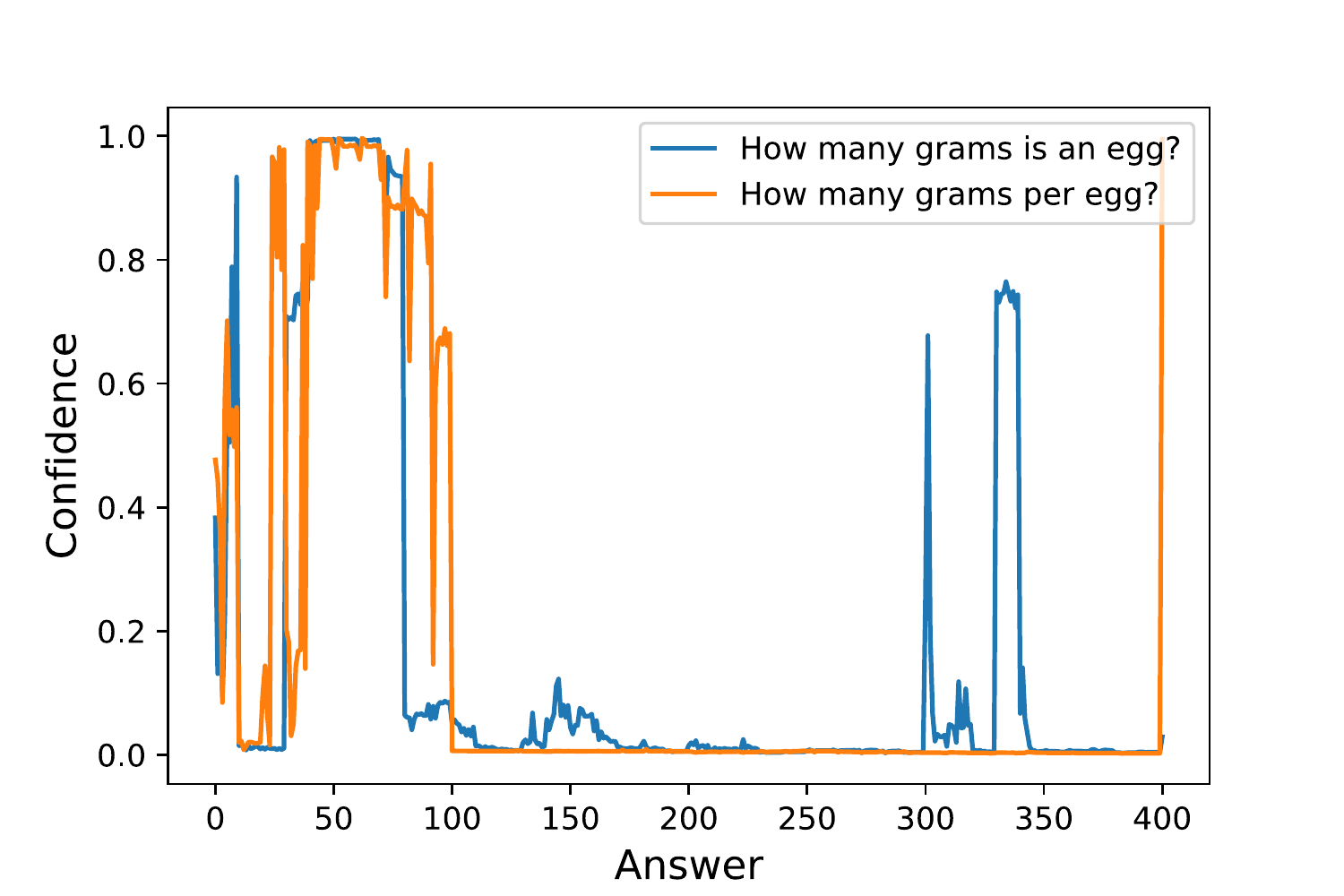}
\caption{The confidence for GPT answering the questions related to the weight of an egg. The fluctuated curve reflects that GPT does not capture the continuous property of numbers.} 
\label{Fig.gpt_conf}
\end{figure}

% Although those results in language understanding are inspiring, some researchers  have also found that there are limitations to such large pre-trained models. 

Despite their inspiring performance in natural language understanding, these pre-trained language models still cannot do consistently well with tasks involving numbers~\cite{thawani2021representing}.
%They cannot do well with tasks involving numbers, which can be interpreted as numeracy. 
For example, as shown in Fig.~\ref{Fig.gpt_conf}, the confidence for GPT answering the questions related to the weight of an egg tends to oscillate with the changes of answers. It exposes that GPT does not fully learn the continuous property of the numbers regarding certain context. 
% \zhc{For example, if the GPT model is trained to answer questions involving numbers, like "How many grams is an egg?",
% it often cannot correctly answer the question and the model behaviors will be strange.
% As shown in~\autoref{Fig.bert_logits},
% XXXXXX} \yong{Zhihua, please paraphrase this sentence.}
% i.e., not continuous in the logits give for the questions and answers pairs (from 0 to 304), as shown in~\autoref{Fig.bert_logits}. 
% \zhihua{Not a good example, may need to replace with GPT example.}
Some researchers~\cite{wallace2019nlp, thawani2021representing} regard it as \textit{numeracy ability}.
% Their research from other perspective also support the conclusions that
Their research has also confirmed that
the large pre-trained model still cannot handle numerical information very well. The lacking of numeracy ability hinders those models from performing well on tasks requiring numeracy~\cite{wallace2019nlp}, which is prevalent in real world problems. 

A line of work has proposed methods to improve the numeracy ability of the neural model. 
%In recent years, there are some research focusing on injecting the numeracy into the model. 
For example, \citet{geva2020injecting} improve the numeracy via generating more training data including numbers. \citet{zhang2020language} transform the number in the text into a scientific notation form. However, it does not guarantee that the models can learn numerical knowledge from the perspective of magnitude and mathematical notation. 
Other studies attempt to design a numeral embedding to improve the numeracy ability. They demonstrate that their designed embeddings can perform well in probing tasks on numeracy~\cite{jiang2019learning, sundararaman2020methods}.
% Embedding trained. -> Incorporate them into the model. 
However, none of them pre-trains their model with numeral embeddings on a large corpus in a self-supervised manner.  Therefore, we want to seek suitable numeral embeddings and incorporate them in pre-training large-scale language models.
% However, they don't investigate incorporating the numeral embedding into the pre-trained language models (e.g., GPT) to improve their numeracy ability. Therefore, we want to seek suitable numeral embeddings and incorporate them in the pre-training stage of models.

Inspired by previous work~\cite{jiang2019learning, sundararaman2020methods, zhang2020language},
% and scientific notation,
we propose NumGPT, a general autoregressive language model that uses a prototype-based embedding to encode the \textit{mantissa} (or \textit{significand}) of the number and separate embeddings to encode the \textit{exponent} of the number. We believe the inductive bias introduced by the numerical embedding will allow the model to present the precision and magnitude of numbers separately and generalize better on the numeric related tasks.  We also design a numeral-aware loss function which enables the model to generate numbers as well as regular text tokens. 
% we propose a numeral embedding based on the scientific notation. 
To evaluate the performance of NumGPT, we synthesize several classification tasks such as measurement estimation, number comparison, and simple arithmetic problems. We also conduct experiments on a real dataset which focuses on magnitude classification, called Numeracy-600K~\cite{chen2019numeracy} to evaluate the numeral predictive ability of our model.
%After obtaining promising results, We further jointly pretrain the GPT model with the numeral embedding on the wikipedia dataset, finetune on MultiNLI and evaluate it on EQUATE dataset. 
Experiments demonstrate that our methods outperform the baseline models including GPT and GPT with DICE~\citep{sundararaman2020methods}. We further evaluate the generated text of NumGPT on a subset of the math word problem dataset, showing that our methods can generate better numerals. Ablation studies on the effect of pre-training and model hyperparameters are conducted to test whether they have a large impact on the model performance.
% We further pre-train the GPT model with the numeral embedding on the Wikipedia dataset and observe whether it can further improve the performance.
% Experiments demonstrate that our methods outperform the baseline models including GPT model in those tasks.
%\yong{Zhihua, please check my side comments.}

Our contributions can be summarized as follows:

\begin{itemize}
  \item We propose a novel NumGPT model, which integrates the specifically designed numeral representations and loss function into the GPT model.
  \item We evaluate the numeracy ability of models on a series of synthetic and real-world tasks, and NumGPT achieves superior performance among them.
\end{itemize}

\begin{figure*}[htb]
\centering 
\includegraphics[width=0.95\textwidth]{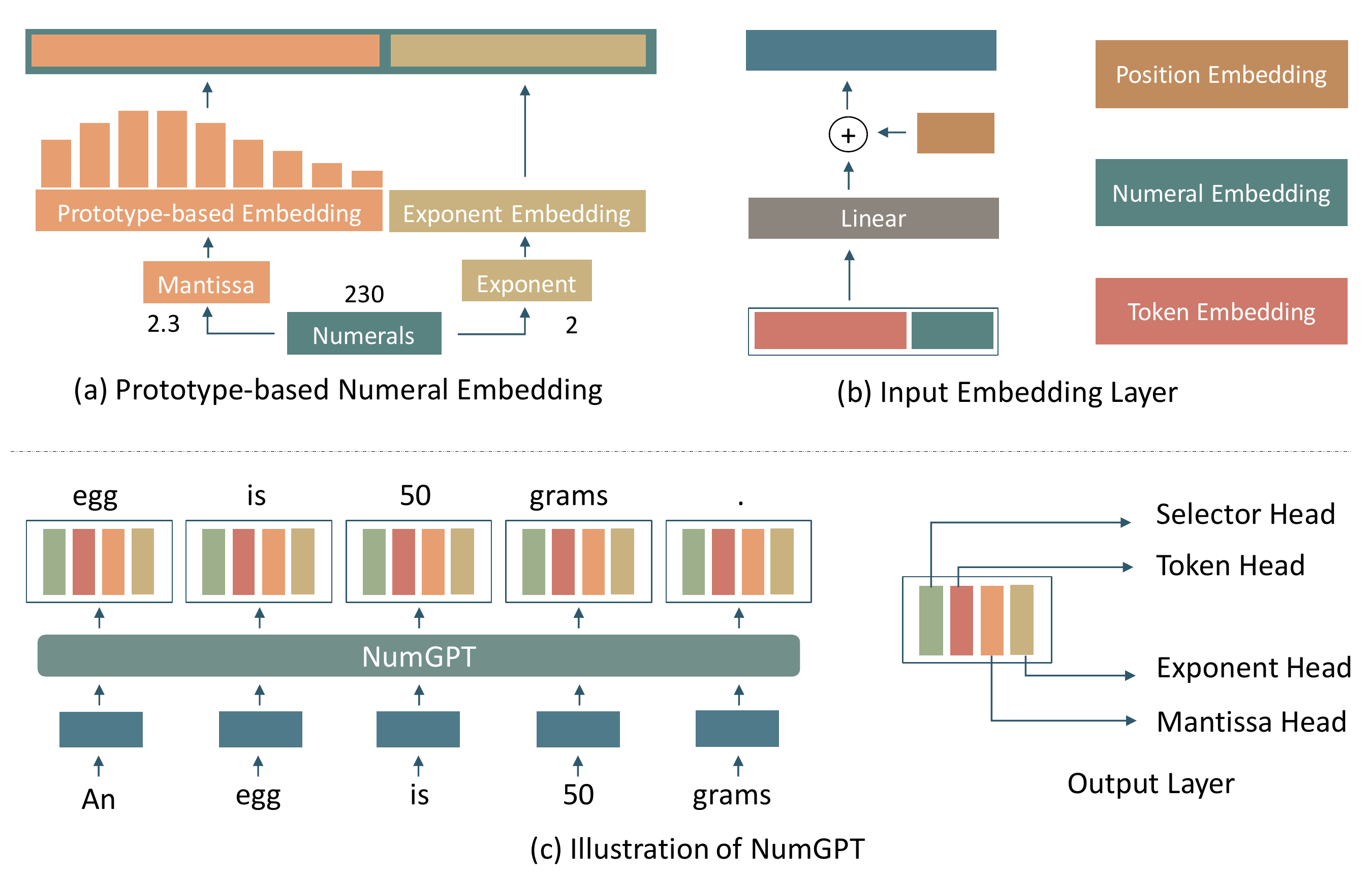}
\caption{The model architecture of NumGPT: (a) We use prototype-based embedding to encode the mantissa of the numeral and use an individual embedding to encode the exponent of the numeral; (b) The numeral embedding and token embedding are fused in the input embedding layer; (c) NumGPT has four heads to control the output. If the Selector Head signals that it is a token, then it will use the Token Head to output a token. Otherwise, it will use Mantissa Head and Exponent Head to output a numeral.} 
\label{Fig.model_arch}
\end{figure*}
\section{Methods}

% 1. Here, we introduce our proposed method NumGPT.
We focus on improving the numeracy ability of GPT in the two stages of the training process, i.e., pre-training and fine-tuning. 
The workflow includes pre-processing the numerals in the corpus, pre-training NumGPT with a numeral-aware language modeling loss, and fine-tuning the model on the domain-specific dataset.
%The first step is to pre-process the numerals in the training corpus. % The numerals need to be extracted and converted to a floating-point number. 
%The second step is to pre-train NumGPT via a numeral-aware language modeling loss. % Since the original language modeling loss only focuses on modeling tokens, we need to adapt it to predict numerals. 
%The last step is to fine-tune the model on the domain-specific dataset. 
In the following sections, we will introduce our methods including numeral parser, numeral embedding, the architecture of NumGPT, and numeral-aware loss function.%, and the decoding process of NumGPT. 
% Here, we introduce our proposed method NumGPT. In the following part we will mainly introduce quantity parser, numeral embedding, model architecture, and loss function.

\subsection{Numeral Parser}
The first step is to pre-process the numerals in text. We design a simple numeral parser to transform the common numerals in the text to the required format. It will be marked and further encoded by numeral embeddings. In real-world data, there exist many kinds of numerals~\cite{ravichander2019equate} and we mainly focus on three typical types:
% quantities: 
number, number with commas, and percentage. For example, "2300", "2,300" should be parsed to 2300. "23\%" should be parsed to 0.23. 
% We use SpaCy to detect entities and post-process the entity including numeral.
%We use SpaCy to detect and post-process entities including numerals.
%The vocabulary of OpenAIGPTTokenizer is adopted to use BPE encoding to encode text.
Besides, we adopt Byte-Pair Encoding (BPE)~\cite{sennrich2015neural} to tokenize the regular text. 
After pre-processing the data, 
the output includes a binary label with each element indicating whether the corresponding element is a (textual) token or a numeral.%, a sequence of tokens, and a sequence of numerals.
% in each position of the sequence, it consists of a selector index which indicates whether it is a token or a numeral. 

% \yong{pls check if it is correct.}
%  there exist some situations that a piece of text can be a numeral and a token, we would like to first simplify the problem via a simple definition.
%and improve the language model to model the joint probability.
%, a token, and a numeral. \jx{(Actually a piece of text could be a numeral and a token, and the language model should be able to model the joint probability. That being said, the selector should not be a hard classifier in principal)}. 
% There exist dummy tokens and dummy numerals to align the position of sequence.  

\subsection{Prototype-based Numeral Embedding}
For GPT, the token IDs will be transformed to embeddings through an embedding matrix. However, in terms of numerals, regular embeddings would not suffice. The reason is that numbers are usually continuous and follow ordered relations, for example, they can be sorted based on their magnitude. In regular GPT, numerals are segmented into one or more independent subword units, which makes it cumbersome to learn the correct relation between numerals. 
Therefore, we need to design specific representations for the numerals so as to model them in an approximately continuous space. 

Recent research has proposed using a deterministic approach~\cite{sundararaman2020methods} or calculating the weighted average of prototype embedding to determine the embedding of numeral~\cite{jiang2019learning}. 
% There is also a research to 
Another work transforms numbers into the form of scientific notation to improve the model capability of capturing scale information~\cite{zhang2020language}. Inspired by these work, we design a hybrid approach to embed the numerals which can capture the scale as well as the precision in a continuous manner.

% Inspired by previous work~\cite{jiang2019learning, sundararaman2020methods, zhang2020language} and scientific notation, we propose a numeral embedding based on the scientific notation.

% After extracting numbers, 
For a numeral $n$,
we will first transform it into a scientific notation and determine its exponent $e(n)\in \mathbb{Z}$ and mantissa $f(n)\in(-10,10)$, as shown in Equation~\ref{eq:sci_transformed}.
\begin{equation}
\label{eq:sci_transformed}
    n= 10^{e(n)} \times   f(n). 
\end{equation}
For example, $-123$ can be transformed to $-1.23 \times 10^{2}$.
% The exponent $e(-123)$ is the $2$. The mantissa $f(-123)$ is $-1.23$.  
Its exponent, denoted as $e(-123)$, is the $2$, and its mantissa, denoted as $f(-123)$, is $-1.23$.  
%\jx{(shall we add an additional bit to indicate if this number is an integer or a decimal?)} 
% A future direction is to record the decimal of the number in the extraction.
Based on the mantissa and exponent of the number, we calculate the numeral embedding $\texttt{NE}(n) \in \mathbb{R}^{d}$, where $d$ is the dimension of numeral embedding.
We encode the mantissa and exponent of the number separately, as shown in Equation~\ref{eq:split_dim} and Fig.~\ref{Fig.model_arch}(a).
\begin{equation}\label{eq:split_dim}
    \texttt{NE}(n)=[\texttt{NE}^{e}(e(n)), \texttt{NE}^{f}(f(n))],
\end{equation}
where $\texttt{NE}^{e}(e(n))\in \mathbb{R}^{d_e}$ is the exponent embedding, and $\texttt{NE}^{f}(f(n))\in \mathbb{R}^{d_f}$ is the mantissa embedding.
$d_f$ and $d_e$ are dimensions of mantissa and exponent embeddings respectively, and $d=d_e+d_f$.
% is , the embedding dimension of exponential is $d_e$,
In the current implementation, we empirically set $d_e=d/4, d_f=3d/4$. %similar to the arrangement of float number representation in computers. 
For the number $-123$, its embedding is $\texttt{NE}(-123)=[\texttt{NE}^{e}(2),\texttt{NE}^{f}(-1.23)]$. We use different strategies to embed mantissa and exponent in the numeral embedding.

% them in the numeral embedding. 
Considering that the exponent is an integer, we associate a learnable embedding vector $\texttt{NE}^e(e(n))\in \mathbb{R}^{d_e}$ to each integer in the typical range of common numbers, that is $\{-8, -7, .., 11, 12\}$. For the exponent larger than $12$ or less than $-8$, we set it to $+\texttt{INF}$ and $-\texttt{INF}$ respectively, as the signs of overflow and underflow for the model. 
% For the mantissa, we pass them into the prototype-based embedding layer to derive the corresponding embedding, which will be introduced in the following paragraphs. 

Since the mantissa is a real number, we choose a different embedding function for it.
Similar to previous work~\cite{sundararaman2020methods,jiang2019learning}, we adopt the deterministic approach and the prototype-based method to define the mantissa embedding. We denote the prototypes as $\{q^f_i\}_{i=0}^{d_f-1}$ where $q_i^f \in [-10, 10]$. Each value of the mantissa embedding is calculated based on the distance between the mantissa and each individual prototype. The formula is shown in Equation~\ref{eq:kernal}.
% The 
\begin{equation}
\label{eq:kernal}
  \texttt{NE}^f_i(f(n))=\exp \left(-\frac{\|f(n)-q^f_i\|^{2}}{\sigma^{2}}\right)  
\end{equation}
where $\sigma$ is a hyperparameter controlling the smoothness of the mantissa representation. In this work, we simply use prototypes uniformly distributed in $[-10,10]$:
\begin{equation}
\label{eq:mantissa_proto}
    q_{i}^{f}=\frac{10-(-10)}{d^{f}-1} \times i + (-10)
\end{equation}

%The typical range of exponential value of the number lies in $[-2,5]$\jx{(could use a larger range), and we learn a distinct embedding for each integer value in it}. \sout{Then we set the min value of exponential prototypes as $-2$ and the max value of exponential prototypes as $5$. The formula for exponential prototypes is defined in~\autoref{eq:exp_proto}.

\subsection{NumGPT}
After deriving the numeral embedding, we can incorporate it into the pre-trained language models, or more specifically, GPT in this work. The architecture of GPT is based on the Transformer decoder. It can be formulated as three modules, i.e., the embedding layer, Transformer layers, and the output layer. The major difference between NumGPT, which is our proposed model, and GPT lies in the input embedding layer and the design of the output layer, as shown in Fig.~\ref{Fig.model_arch}(b) and Fig.~\ref{Fig.model_arch}(c) respectively.

\noindent\textbf{Input Embedding Layer.} 
In the design of the original Transformer~\cite{vaswani2017attention}, it derives the input embedding through adding the token embedding and position embedding. Then it will be passed to the Transformer layer $h=\texttt{Transformer}(\texttt{E}(x_t))\in \mathbb{R}^{d_h}$, where $d_h$ is the hidden size and $h$ is the hidden state. 
Consider the token embedding function as $\texttt{TE}(x_t) \in \mathbb{R}^{d_h}$ and the position embedding function as $\texttt{PE}(x_t) \in \mathbb{R}^{d_h}$, the output of input embedding layer $\texttt{E}(x_t) \in \mathbb{R}^{d_h}$ is shown in Equation~\ref{eq:ori_input_emb}:
\begin{equation}
\label{eq:ori_input_emb}
    \texttt{E}(x_t)=\texttt{TE}(x_t) + \texttt{PE}(x_t),
\end{equation}
where $t \in \{1, 2, .., T\}$ is the position in the sequence $x$ and $T$ is the length of sequence. % and $d_h$ is the size of hidden states. 

% Since we have introduced the numeral embedding for numbers, we need to 
% this new information. 
For NumGPT, we modify the input embedding layer to incorporate the numeral embedding for numbers.
A paradigm is to concatenate the token embedding and the numeral embedding and then use a linear layer to transform it into the same size of hidden states. 
The formula is shown in Equation~\ref{eq:new_input_emb}.
\begin{align}\label{eq:new_input_emb}
    \texttt{E}(x_t) &= [\texttt{TE}(x_t), \texttt{NE}(x_t)]W_b + \texttt{PE}(x_t),
\end{align}
where $W_b \in \mathbb{R}^{(d_h+d) \times d_h}$ denote parameters of the linear transformation.
% Since the token and numeral are complementary, we design an indicator $Sel(seq_t) \in \{0, 1\}$ to denote whether it is a token or numeral, 
% where $0$ represents that the position $t$ of the sequence is a token and otherwise a numeral.
%and $1$ represents that it is a numeral. 
% According to $Sel(seq_t)$, 
% we can select which part is actively used. 
\zhihua{Since the token $x_t$ in the sequence can be textual or numeral exclusively, when the token 
is textual, its numeral part $\texttt{NE}(x_t)$ becomes a learnable embedding shared among all the textual tokens. When the token 
is a numeral, the textual part $\texttt{TE}(x_t)$ for the token becomes a learnable embedding shared across all the numerals. }

\noindent\textbf{Output Layer.} 
In general, the output probability of the next token follows a mixture model: 
\begin{align}
    p(x_t|&x_{1:t-1})  \nonumber\\
    = & p(z_t=1|x_{1:t-1})p(x_t|x_{1:t-1},z_t=1) \nonumber\\
    &+ p(z_t=0|x_{1:t-1})p(x_t|x_{1:t-1},z_t=0), 
\end{align}
% with the variable $z_t=0$ for regular token and $z_t=1$ for numeral. 
where $z_t=0$ indicates a textual token and $z_t=1$ represents a numeral token. In the output layer, we decouple each part and model them separately.
First, we use Selector Head to output the probability of token types. It can be calculated as follows:
\begin{equation}
    p(z_t|x_{1:t-1}) =\texttt{softmax}(h_tW_z).
\end{equation}
where $h_t\in \mathbb{R}^{d_h}$ is output hidden state and $W_z\in \mathbb{R}^{d_h\times 2}$ is the parameter matrix.

% The design of output heads in NumGPT is different from that of the original GPT.
% when pre-training GPT. 
Then, if the Selector Head predicts the next token as textual, the Token Head will output the probability of the next tokens $\texttt{P}(h) \in \mathbb{R}^{T\times |V|}$, as shown in Equation~\ref{eq:ori_predtok}.
\begin{equation}\label{eq:ori_predtok}
   p(x_t|x_{1:t-1},z=0) =\texttt{softmax}(h_tW),
\end{equation}
where $W\in \mathbb{R}^{d_h\times |V|}$ is the parameter matrix, and $|V|$ is the size of vocabulary. 

For a numeral token $x_t$, we decompose it into two parts $[x^e_t,x^f_t]$, which correspond to exponent and mantissa.
They are regarded as independent variables and their marginal probabilities are modeled as:
% for three parts of sign, exponent and mantissa as independent variables and model their marginal probabilities as:
 %   p(x^s_t|x_{1:t-1}, z_t=1) &= \texttt{softmax}(h_tW_{s}) \\
\begin{align}
    p(x^e_t|x_{1:t-1}, z_t=1) &= \texttt{softmax}(h_tW_{e}) \\
    p(x^f_t|x_{1:t-1}, z_t=1) & = \frac{1}{\sqrt{\pi}}\exp(-(x^f_t- h_tW_{f})^2)
\end{align}
where $W_e\in \mathbb{R}^{d_h\times |V_e|}$ and $W_f\in \mathbb{R}^{d_h\times 1}$ are parameter matrices, and $|V_e|$ is the size of exponent vocabulary. $h_tW_{f}$ can be regarded as the mantissa of the predicted numeral.
    %p(x^f_t|x_{1:t-1}, z_t=1) & = \frac{1}{Z}\exp(-(x^f_t*10^{x^e_t}- h_tW_{f}*10^{x^e_t})^2),
% where $Z$ is a normalization constant. 
% where $g$ is a normalized function measuring the similarity between the continuous numeral $x_t^f$ and the predicted probability $\texttt{softmax}(h_tW_{f})$ by the model. %Moreover, $W_{sel}\in \mathbb{R}^{d\times 2}$, $W_{sign}\in \mathbb{R}^{d\times 2}$, $W_{frac}\in \mathbb{R}^{d\times d_f}$, $W_{exp}\in \mathbb{R}^{d\times d_e}$ are learnable parameter matrices. We will show how we decode the numeral from the above heads in Section \ref{sec:decode_numgpt}. 

\subsection{Numeral-aware Loss Function}
Finally, we obtain a numeral-aware language modeling loss function to pre-train the model: %and fine-tune a model using cross entropy loss. The loss function of NumGPT is to maximize the probability of the generated sentences $p(x_1,x_2,...,x_T)$, which is defined as follows:
\begin{align}
    %\mathcal{L} = - &\sum_{t=1}^{T} \log p(x_t|x_{1:t-1}) \nonumber\\
    \mathcal{L} = - &\sum_{t=1}^{T} [\log p(z_t|x_{1:t-1}) \nonumber\\
    & + I(z_t=0) \log p(x_t|x_{1:t-1}, z_t=0) \nonumber\\
    & + I(z_t=1) \log p(x^e_t|x_{1:t-1}, z_t=1) \nonumber\\
    & + I(z_t=1) \log p(x^f_t|x_{1:t-1}, z_t=1) ] .
\end{align}

% & + I(z_t=1) \log p(x^s_t|x_{1:t-1}, z_t=1) \nonumber\\
\section{Experiments}
% 1. Here, we introduce experiment results for demonstrating our performance.
To evaluate the effectiveness of 
% our method 
NumGPT in understanding numeracy, we conduct the following experiments. 
Inspired by previous evaluation methods~\cite{hosseini2014learning,talmor2019olmpics, zhang2020language}, we first synthesize three datasets as \textbf{Synthetic Tasks} to evaluate the numeracy ability of models.
%several challenging tasks,
% to evaluate our proposed model, 
%which is so-called \textbf{Synthetic Tasks}. 
We further demonstrate the application of NumGPT in the generation task and magnitude classification task~\cite{chen2019numeracy}. 
Ablation studies are designed to investigate whether pre-training and the hyperparameters of NumGPT have a great impact on the performance of models. %and which hyperparameters are more important.
% \yong{Zhihua, please check my Chinese comments.}

\subsection{Synthetic Tasks}
% To fully expose the drawbacks of GPT on numeracy, we design three synthetic tasks. 

% They are 
We first compared NumGPT with baseline approaches, including GPT, GPT with DICE~\citep{sundararaman2020methods}, on three synthetic tasks:
Multiple Measurement Estimation (MME) task, General Number Comparison (GNC) task, and Math Word Problem Addition and Subtraction (MWPAS) task. 
%\yong{Please introduce the baseline approaches as well.}
We create tens of templates for the tasks and sample random numbers to fill in the template. 
We frame those tasks as a binary classification problem, which
% is judging 
judges
whether the question-answer pair or the sentence is "correct" or not. 
We evaluate the model by calculating accuracy in the test dataset. We illustrate them one by one in the following paragraphs.

\noindent\textbf{Multiple Measurement Estimation.} Inspired by the measurement estimation task proposed by~\citet{zhang2020language}, we design the MME task to verify whether models can learn to estimate objects' numerical attributes. % There are also similar measurement estimation tasks in. 
We construct a question-answer pair and let the model judge whether the answer matches the question. 
One question template is "How many grams are [INT] [OBJ]?" and we will sample a random integer to fill in "[INT]" and an object, such as "egg", to fill in "[OBJ]". We assume that the range of "correct" answer is $[\texttt{ANS\_MIN}, \texttt{ANS\_MAX}]$. We limit the range of "incorrect" answers in $[0.01\times \texttt{ANS\_MIN}, \texttt{ANS\_MIN})\cup (\texttt{ANS\_MAX}, 100\times \texttt{ANS\_MAX}]$. For example, a question is "How many grams are 2 eggs?" and the answer range for this question is $[70, 140]$. If we input this question with the answer "35" to the model, the model should classify this question-answer pair as "incorrect". If the input answer is changed to "80", the model should classify this question-answer pair as "correct".
%We consider that if the model predicts that one egg weigh 35 grams to 75 grams, then it is considered "correct", otherwise it is considered "wrong". 
%Since the question asks the grams of 2 eggs, then the "correct" answer range lies in $[70, 140]$. If the answer $150$ is passed into the model, the model should predict that it is a "wrong" answer. If the answer is $78$, then the model should predict that it is a "correct" answer. 
For the dataset of this task, we have crafted 20 objects and 4 question templates. When generating one sample, we randomly sample a question template from candidate question templates, use the logarithmic sampler to sample the multiplier, and sample a candidate answer. For each object, we construct 500 "correct" samples and 500 "incorrect" samples. Then we split the generated samples into 16000 training samples and 4000 test samples. We guarantee that the combination of multipliers, object, and answer is unique for all the samples in the dataset.

\noindent\textbf{General Number Comparison.} Inspired by the number comparison task proposed in oLMpics~\cite{talmor2019olmpics}, we augment it as general number comparison to evaluate whether the model can judge the quantitative comparison in the natural language context. 
%This task is similar to the number comparison task in the oLMpics~\cite{talmor2019olmpics}.
A template for this problem is "A [NUMA] year old person is younger than a [NUMB] year old person", where "[NUMA]" and "[NUMB]" can be replaced with numbers in a range $[15, 104]$. The label is determined based on whether two numbers satisfy the semantics of the sentence. We have crafted 20 templates, which cover a large range of numbers and typical object numerical attributes (e.g., length and weight) comparison, and then according to each template, we generate 1000 positive samples and 1000 negative samples. In each sample, the number is randomly sampled using a logarithmic sampler to fill in the template. The dataset can be split into 32000 training samples and 8000 test samples. %Other settings are the same as the MME task.

\noindent\textbf{Math Word Problem Addition and Subtraction.} The MWPAS task assesses whether the model can handle the addition and subtraction task in the math word problem. Twenty templates are crafted from AI2 dataset~\cite{hosseini2014learning}. They cover the ability of addition and subtraction. 
For example, one template is "Joan found [NUMA] seashells on the beach. She gave Sam some of her seashells. She has [NUMB] seashells. How many seashells did she give to Sam?". Similarly, the "[NUMA]" and "[NUMB]" can be replaced by random integers. If the answer equals the subtraction from Number A to Number B, then the label is "true". Otherwise, it is "false".
% Similar to the settings of generation in the GNC task, We have crafted 20 templates, which cover the ability of addition and subtraction and a large range of answers. 
For each template, 1000 positive samples and 1000 negative samples are created. In each sample, numbers are randomly sampled using a logarithmic sampler to fill in the template. The splitting strategy is the same as the GNC task.%The dataset can be split into 32000 training samples and 8000 test samples.%We follow the same settings in the GNC task.

\begin{table}[]
\centering
\small
\begin{tabular}{lccc}
\hline
Model & MME & GNC & MWPAS \\ \hline
MAJ & 50.48$\pm$0.00 & 50.93$\pm$0.00  & 50.93$\pm$0.00 \\
 &  &  &  \\
\multicolumn{4}{l}{\textit{Train   from scratch}} \\
GPT & 78.99$\pm$1.81 & 95.16$\pm$0.28 & 49.85$\pm$0.63 \\
GPT with DICE & 73.68$\pm$0.43 & 75.09$\pm$1.21 & 49.17$\pm$1.04 \\
NumGPT & 97.16$\pm$0.55 & 95.45$\pm$0.57 & \textbf{88.05}$\pm$0.99 \\
 &  &  &  \\
\multicolumn{4}{l}{\textit{Pre-train   and finetune}}  \\
GPT & 72.02$\pm$3.63 & 93.84$\pm$1.29 & 49.17$\pm$0.82 \\
NumGPT & \textbf{98.11}$\pm$0.38 & \textbf{95.82}$\pm$0.17 & 86.16$\pm$3.26 \\ \hline
\end{tabular}

\caption{Experiment results for synthetic tasks. \zhihua{The performance of NumGPT is better than baseline models in the three synthetic tasks \pmark. The pre-training improves the performance of NumGPT on MME task \pmark. }}
\label{tab:syn_acc}
\end{table}

\noindent\textbf{Model Details.} In this study, we investigate the model performance on synthetic tasks. Baseline models include majority baseline (MAJ), GPT, and GPT with DICE. The MAJ baseline simply selects the most frequent label in the test dataset as the output. GPT is a Transformer decoder with 12 layers, 12 heads, and the hidden size of 768. NumGPT has the same architecture as GPT. The numeral embedding dimension of NumGPT is 64 and $\sigma$ is 0.5. For GPT with DICE, we replace the numeral embedding in NumGPT with DICE~\cite{sundararaman2020methods}. \zhihua{We train the models from scratch for 50 epochs on one Nvidia V100 GPU with batch size of 96.} The optimizer we used is AdamW and the learning rate is $6.25\times10^{-5}$. 

% \yong{Reach here.}

\noindent\textbf{Results.} The results of experiments on synthetic tasks are listed in Table~\ref{tab:syn_acc}. \zhihua{The mean and standard derivation of accuracy over 5 runs are reported. We also performed the Student's $t$-test to determine whether the average performance scores of two groups are significantly different. We assume that if the $p$-value is less than 0.05, the difference is statistically significant. %The conclusion with p-value less than 0.05 will be reported. 
NumGPT outperforms all the baseline methods in MME and MWPAS tasks \pmark~and achieves the comparable results of GPT in GNC task. Those tasks focus on evaluating different aspects of the numeracy ability.  
For the MME task, we find that NumGPT can achieve a significant improvement compared to GPT and GPT with DICE \pmark.} It demonstrates that our numeral embedding captures the scale information more accurately, since it models the exponent individually. It further facilitates measurement estimation with multiplication, while GPT and GPT with DICE do not perform very well on it.
% cannot learn such ability. 
\zhihua{For the GNC task, NumGPT has a slightly better performance than GPT.} However, GPT with DICE cannot achieve a good performance on this task. One possible reason is that when integrating the DICE embedding into the model, the model cannot capture the tiny difference in the embedding if the numbers are very close. 
Therefore, it is hard for the model to distinguish similar numbers, which leads to a lower performance. For the MWPAS task, GPT and GPT with DICE have a similar performance as the MAJ baseline model. 
It demonstrates that they cannot model the addition and subtraction in the math word problems. \zhihua{NumGPT significantly outperforms other baseline models in this task \pmark.} It reflects that NumGPT has a more accurate arithmetic ability than other baseline models. A reason is that the numeral embedding of NumGPT eases the difficulty of modeling addition and subtraction.   % It has a large amount of improvement compared to GPT and GPT with DICE on MME and MWPAS tasks.
% Therefore, we draw a conclusion that the numeracy ability of NumGPT is greatly improved. 

\subsection{Magnitude Classification}
Magnitude classification on \textbf{Numeracy-600K} dataset~\cite{chen2019numeracy} is a task requiring models to predict the magnitude of the masked numeral in the news titles. We conduct the experiments on this dataset and compare the performance of our proposed model with that of the baselines. 

\noindent\textbf{Model Details.} \zhihua{GPT is a Transformer decoder with 12 layers, 12 heads, and the hidden size of 768. NumGPT has the same hyperparameter configuration as GPT. We include the results of GPT and NumGPT. They are trained from scratch on one Nvidia V100 GPU with a batch size of 96 for 5 epochs. The optimizer used in this experiment is AdamW and the learning rate is set as $6.25\times10^{-5}$. }

\noindent\textbf{Results.} 
% Results can be found in~\autoref{tab:num_acc}. 
Table~\ref{tab:num_acc} shows the experiment results on magnitude classification. \zhihua{Average values and standard deviations of Micro-F1 and Macro-F1 over 5 runs for GPT and NumGPT are reported respectively.}
Compared to the baseline results provided by~\citet{sundararaman2020methods}, we can observe that our approach outperforms all the baseline models.
\zhihua{Also, our model achieves a comparable performance with GPT. It indicates that NumGPT has a similar numeracy ability for magnitude prediction with GPT when training from scratch. A possible reason is that many numerals are masked in the input and NumGPT cannot show its representation power of numerical embeddings by only learning from the task dataset. While after pre-training the NumGPT on a large unlabeled corpus, it can achieve better performance than GPT on this task \pmark. It demonstrates that through fully learning numeral relations in pre-training dataset, the performance of NumGPT can be improved on the downstream tasks.} 
% It demonstrates that our model can also capture numeracy in this study.
\begin{table}[]
\centering
\begin{tabular}{lll}
\hline
Model & \multicolumn{1}{l}{Micro-F1~$\uparrow$} & \multicolumn{1}{l}{Macro-F1~$\uparrow$} \\ \hline
LR & 62.49 & 30.81 \\
CNN & 69.27 & 35.96 \\
GRU & 70.92 & 38.43 \\
BiGRU & 71.49 & 39.94 \\
CRNN & 69.50 & 36.15 \\
CNN-capsule & 63.11 & 29.41 \\
GRU-capsule & 70.73 & 33.57 \\
BiGRU-capsule & 71.49 & 34.18 \\
BiLSTM with DICE & 75.56 & 46.80 \\ \hline
\multicolumn{3}{l}{\textit{Train from scratch}} \\
GPT & 79.45$\pm$0.45 & 53.86$\pm$0.84 \\
NumGPT & 78.97$\pm$0.24 & 53.30$\pm$1.12 \\
 &  &  \\
\multicolumn{3}{l}{\textit{Pre-train and finetune}} \\
GPT & 79.42$\pm$0.15 & 53.79$\pm$0.68 \\
NumGPT & \textbf{81.32}$\pm$0.21 & \textbf{56.47}$\pm$0.77 \\ \hline
\end{tabular}
\caption{Experiment results on Numeracy-600K dataset. \zhihua{NumGPT outperforms other baseline models \pmark}.}
\label{tab:num_acc}
\end{table}

% The baseline results refer to~\citet{sundararaman2020methods}. 

\subsection{Generation Evaluation}
% In order to 
To further evaluate the quality of generation task, we further conduct the generation evaluation based on the MWPAS task. We construct the input context based on the positive samples with answers larger than 10000 in the MWPAS task. 
% For example, one input context is "Q: There were 97 roses in the vase . Mary cut some roses from her flower garden . There are now 98 roses in the vase . How many roses did she cut? A:" and then predict the next token, which is expected to 1.
For example, one input context is "Q: A ship is filled with 6518 tons of cargo. It stops in Bahamas, where sailors load 3542 tons of cargo onboard. How many tons of cargo does the ship hold now? A:" and the model will predict the next word, which is expected to "10060". 
We split the dataset into 5512 training samples and 1338 test samples. The model details, evaluation metrics, and results will be introduced in the following paragraphs.

\noindent\textbf{Model Details.} We train GPT and NumGPT using language modeling loss in this task. Similar to the architecture setting of Synthetic Tasks, GPT is a Transformer decoder with 12 layers, 12 heads, and the hidden size of 768. NumGPT has the same hyperparameter configuration as GPT. \zhihua{They are trained from scratch on one Nvidia V100 with batch size 96 for 50 epochs.} The optimizer used in this experiment is AdamW and the learning rate is set as $6.25\times10^{-5}$.

\noindent\textbf{Evaluation Metrics.} We find that it is also possible for the model to predict a token other than a number. Therefore, we define several metrics to evaluate how well the model performs in this task.
% For GPT, it is very likely to generate a token other than number. Therefore, we define several metrics to evaluate how well the model performs in this task. 
The first metric is the Number Generation Ratio ($\texttt{NGR}$), which is a value from 0 to 1. It measures how many percentages of samples will lead to
% the model predict the numbers. 
the final numerical predictions by the model.
We use the next two metrics log-MAE ($\texttt{LMAE}$) and exponent accuracy ($\texttt{E\_Acc}$) with log base 10~\cite{berg2020empirical} to measure the correctness of generated numbers, which are defined as follows:
\begin{align}
\texttt{LMAE}&=\frac{1}{\left|\mathcal{D}\right|} \sum_{\mathcal{D}}|\log y-\log \hat{y}| \\
\texttt{E\_Acc}&=\frac{1}{\left|\mathcal{D}\right|} \sum_{\mathcal{D}} \mathbb{I}[\lfloor\log y\rfloor=\lfloor\log \hat{y}\rfloor]
\end{align}
where $\mathcal{D}$ is the sample collection, $y$ is the ground truth, and $\hat{y}$ is the model prediction. 

\begin{table}[]
\centering
\begin{tabular}{lccc}
\hline
Model & NGR & LMAE~$\downarrow$ & E\_Acc~$\uparrow$ \\ \hline
\multicolumn{4}{l}{\textit{Train from scratch}} \\
GPT & 1$\pm$0 & 0.0957$\pm$0.0045 & 0.9221$\pm$0.0030 \\
NumGPT & 1$\pm$0 & \textbf{0.0312}$\pm$0.0044 & \textbf{0.9749}$\pm$0.0188 \\
 &  &  &  \\
\multicolumn{4}{l}{\textit{Pre-train and finetune}} \\
GPT & 1$\pm$0 & 0.0923$\pm$0.0084 & 0.9202$\pm$0.0117 \\
NumGPT & 1$\pm$0 & 0.0371$\pm$0.0099 & 0.9430$\pm$0.0728 \\ \hline
\end{tabular}
\caption{Experiment results on generation evaluation on a subset of the MWPAS dataset. \zhihua{NumGPT outperforms GPT in this task \pmark}.}
\label{tab:gen_eval}
\end{table}

\noindent\textbf{Results.} 
The experiment results are shown in Table~\ref{tab:gen_eval}. \zhihua{The mean and  standard deviation of performances over 5 runs are reported. }
Although the models have the ability to generate a token rather than a numeral, we find that the $\texttt{NGR}$ for both models is 1, which means that both models can generate a number given the input context. We can also observe that NumGPT outperforms GPT in terms of \texttt{LMAE} and \texttt{E\_ACC}. It demonstrates that NumGPT is more accurate in generating large numbers than GPT.
%It demonstrates that the generated numbers of NumGPT are more reasonable than GPT.  %It demonstrates that the pre-training stage can capture additional numeracy containing in the dataset.

% \yong{reach here.}
\subsection{Ablation Study}
\label{sec:ablation_study}
We want to identify whether the hyperparameters $\sigma$ used in the mantissa embedding
will affect the model performance. Also, we would like to investigate whether the pre-training will have an impact on the model performance. \zhihua{We conduct both an ablation study regarding $\sigma$ on the MME task and another ablation study regarding pre-training on the synthetic tasks, magnitude classification, and generation evaluation.}
\begin{figure}[htb]
\centering 
\includegraphics[width=0.47\textwidth]{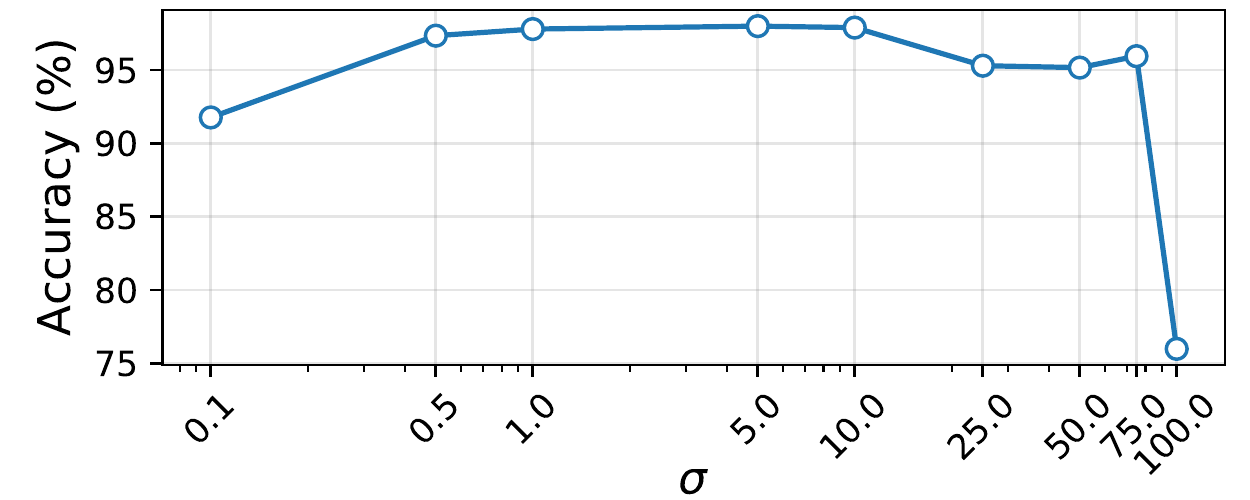}
\caption{Experiment results for the ablation study on $\sigma$. \zhihua{The performance of NumGPT on the MME task will maintain in a small range except when $\sigma$ is very small, such as $0.1$, or $\sigma$ is very large, such as $100$.}} 
\label{fig:SA_sigma}
\end{figure}

\noindent\textbf{Ablation on $\sigma$.} \zhihua{We train NumGPT with $\sigma$ varying from $0.1$ to $100$. Other training settings are the same as the previous experiments on MME tasks. 
%We modify the $\sigma$ value from $0.1$ to $9$ and plot a figure to demonstrate the effect, 
As shown in Fig.~\ref{fig:SA_sigma}, we can observe that as the $\sigma$ increases, the performance is increasing and then decreasing after $\sigma$ is larger than 10. It seems that a very small $\sigma$ value, such as 0.1, or a very large $\sigma$ value, such as 100, will have a significant bad effect on the model performance. The reason is that when the value of $\sigma$ is very small, the numeral embedding will become sparse and then provide not enough detailed information to depict the numerals. When the value of $\sigma$ is very large, the numeral embedding will become very smooth and will not be able to distinguish between two mantissas.
Therefore, we should choose a suitable $\sigma$ to help the model understand the mantissa of the numeral.}

\noindent\textbf{Ablation on Pre-training.} \zhihua{For the pre-training, we pre-train GPT and NumGPT on the Wikipedia dataset, which contains 2,500M words, for one epoch on 8 Nvidia V100 GPUs with batch size 80. For the fine-tuning stage, we train the same epochs as training from scratch on the corresponding training dataset on one Nvidia V100 GPU with batch size 96.}
\zhihua{The experiment results are shown in Table~\ref{tab:syn_acc}, Table~\ref{tab:num_acc}, and Table~\ref{tab:gen_eval}. We repeatedly run the finetuning given the pre-trained weight in 5 runs and report the average and standard deviations of performance scores. The pre-trained NumGPT will improve the performance on MME task, and magnitude classification task \pmark. After pre-training the models, in the magnitude classification task, NumGPT achieves more significant improvements than GPT \pmark. It indicates that NumGPT can achieve better numeracy ability through pre-training on the large unlabeled corpus. While in GNC task, MWPAS task, and generation evaluation task, the performance improvements through pretraining on NumGPT are not significant. One possible reason is that the Wikipedia dataset does not have many samples requiring the arithmetic ability, such as numerical addition and subtraction required in the downstream tasks. 
% However, in MWPAS task, pre-training does not significantly improve the performance of NumGPT. One possible reason is that the Wikipedia dataset does not have many samples requiring the ability of numerical addition and subtraction. 
We also find that pre-training cannot improve performance of GPT in synthetic tasks, magnitude classification, and generation evaluation. It reflects that it is hard for GPT to learn numeracy skills through pre-training in a large unlabeled corpus.}

%We design several templates to generate the samples in the training dataset and test dataset. For measurement estimation, we focus on whether the model can judge a certain range of object attributes are reasonable. Therefore, we design a two class classification problem to evaluate the models ability.
%\subsection{Number Comparison Task}
%Similar to the previous task, we design several template to reflect the number comparison ability on it.

%\subsection{Arithmetic Math Word Problem Task}

%\subsection{Pretrained NumGPT on Wikipedia, Finetune on MultiNLI and evaluate on EQUATE}

%\input{sections/5-discussion}
\section{Related Work}
% 1. We summarize related work into three catogies: numerical understanding tasks, models, and datasets.
% We summarize related work into three categories: pre-trained models for NLP, probing numeracy in NLP models, and quantitative reasoning.
We summarize related work into two categories: probing numeracy in NLP models and methods for improving numeracy ability of NLP models.

\noindent\textbf{Probing Numeracy in NLP Models.} Numeracy ability is mainly about reasoning with numbers in the text. The tasks for probing numeracy ability in NLP models can be further classified into two classes, approximate and exact, depending on the encoding of the numbers in text~\cite{thawani2021representing}. For example, "50" in the sentence "An egg weighs 50 grams." is an approximation of an egg weight,
% approximate number for the weight of an egg may vary in a range. 
while the number "7" in the sentence "3 balls + 4 balls = 7 balls." is an exact answer.
% number for it can be derived from the sentence. 
The probing tasks on numeracy for approximate numbers consist of
% For probing tasks on numeracy ability about approximate numbers, they consist of
numeration~\cite{naik2019exploring, wallace2019nlp}, magnitude classification~\cite{chen2019numeracy}, and measurement estimation~\cite{zhang2020language}.
%\citet{chen2019numeracy} proposed Numeracy600K dataset to evaluate numeracy ability of models.
%\citet{naik2019exploring} build up several kinds of methods to evaluate numeracy of embedding. \citet{zhang2020language} proposed evaluating the scale information of embedding.
The probing tasks on numeracy for exact numbers include number comparison~\cite{talmor2019olmpics} and math word problems~\cite{ravichander2019equate}. 
%a number comparison task is proposed in the oLMpics~\cite{talmor2019olmpics} to investigate the numeracy ability between BERT and RoBERTa. \citet{wallace2019nlp} proposed several probing tasks to evaluate numeracy ability of models. \citet{ravichander2019equate} proposed EQUATE dataset to evaluate the model the ability of natural language inference with numeral.
In this paper, similar to the previous tasks, we synthesize the measurement estimation task, the number comparison task, and math word problems. Besides, we adopt magnitude classification tasks~\cite{chen2019numeracy} to comprehensively evaluate the numeracy ability of models.
% Some researchers have focused on detecting whether there exist numeracy in NLP models and they design a series of probing tasks to reveal it. For example, 
% \citet{spithourakis2018numeracy} proposed several methods to better model numerals. 

%Order: oLMpics -> Numeracy for language models -> Numeracy 600K -> Exploring Numeracy in word embedding -> Do NLP Models know number? -> -> Numeral Embedding -> Learn scale? 

\noindent\textbf{Methods for Improving Numeracy Ability of NLP Models.} Researchers have explored some methods to improve the numeracy ability of NLP models. 
%For methods on numeracy ability about approximate numbers, 
%they mainly focus on designing better embedding to fully learn the numeracy ability~\cite{jiang2019learning,sundararaman2020methods}. 
A line of work focuses on developing a domain-specific problem solver integrated with neural networks and symbolic functions for math word problems~\cite{zhang2019gap}.
% improving numeracy ability about exact numbers. 
% The ability for quantitative reasoning for NLP models is strongly desired for a long time in the field of AI and NLP. However, achieving such ability is not a trivial work.
Their methods are mainly based on expression tree~\cite{roy2016solving}, sequence to sequence model~\cite{wang2017deep}, reinforcement learning~\cite{huang2018neural, wang2018mathdqn}, and hybrid model~\cite{amini2019mathqa,chiang2018semantically, griffith2019solving}. 
% There are several survey on math word problem. A series of papers are focusing on improving model ability on math word problem~\cite{roy2016solving,wang2017deep,huang2018neural,wang2018mathdqn,amini2019mathqa,chiang2018semantically,griffith2019solving}.  
% Another task is to integrate numerical reasoning into the reading comprehension task, which is called DROP~\cite{dua2019drop}. The methods for improving the performance of this task are mainly based on combining symbolic functions~\cite{andor2019giving}, plugging additional modules like graph neural networks to solve specific tasks~\cite{ran2019numnet,gupta2019neural,chen2020question}, or adding more numerical training data during the pre-training stage~\cite{geva2020injecting}. 
%Recent researcher proposed DROP to further investigate the ability of pretrained language models. 
% Different from developing domain-specific problem solvers, a 
Another line of work focuses on improving numeracy in word embeddings or pre-trained models~\cite{jiang2019learning,sundararaman2020methods, berg2020empirical}, which can be generalized well across different tasks. Specifically, \citet{jiang2019learning} proposed learning a weighted prototype numeral embedding and demonstrated that it can perform well on numeral prediction and sequence labeling tasks. However, the training process of the numeral embedding is not integrated into the training process of the models, which leads to suboptimal and time-consuming. \citet{sundararaman2020methods} designed a deterministic corpus-independent numeral embedding 
with excellent performance on probing tasks for numeracy~\cite{naik2019exploring, wallace2019nlp}.
% and demonstrate that it has excellent performance on probing tasks for numeracy~\cite{naik2019exploring, wallace2019nlp}. 
However, when range of numbers becomes large, the embeddings for similar numbers will be very close and hard to distinguish. Moreover, in some tasks like math word problems, a subtle change in the number leads to totally different answers.
While \citet{berg2020empirical} conducted an empirical analysis of different loss functions in the task of contextualized numeral predictions in BERT,
our work explores how GPT can benefit from numeral-aware loss function, similar to their designed loss function. Particularly,
% \citet{berg2020empirical} conducted an empirical analysis of different loss functions in the task of contextualized numeral predictions in BERT. It leaves the space for exploring how GPT can benefit from their designed loss function. Our work falls in the latter category. Inspired by the previous approaches and noticing the limitations of them, 
we propose a deterministic numeral embedding and further integrate it in the pre-trained model GPT. Such a deterministic embedding considering the scientific notation of the numbers is more robust than the numeral embedding proposed by \citet{sundararaman2020methods} for  it is more scalable to produce embeddings for large numbers.

%Order: Survey Math Word Problem -> Math Word Problem Model --> EQUATE -> DROP -> Various models.
\section{Conclusion}
% 1. We proposed the method NumGPT. 
To 
% further 
improve the numeracy ability of GPT, we have proposed NumGPT, which incorporates the numeral embedding into the input embedding. For the numeral embedding, we used scientific notation to decompose the number into mantissa and exponent and embed them into separate parts. Moreover, we designed a numeral-aware loss to handle the generation of numeral.
%The training loss is slightly modified and decoding process has slightly difference. 
% 2. We conducted the experiments to demonstrate XXX.
We have conducted the experiments to demonstrate the effectiveness of our method in the synthetic tasks, magnitude classification, and generation evaluation. We also conduct a series of ablation studies to test whether the hyperparameters and pre-training have a large impact on model performance. From the experiment results, the performance improvement encourages us that integrating numeral embedding into the GPT is a promising direction to improve the numeracy ability of language models. 

In the future, we plan to extend our methods to a larger size model and conduct quantitative experiments on more numerical reasoning tasks. 
Also,
% the explanation of numerical capability by large-scale GPT model is still undiscovered.
the explanation of numerical capabilities of large-scale language models are still less explored.
It will be interesting and valuable to further conduct research in future work.

% could be valuable and offer more insights to further research.
% Also, we would like to further explore whether other numeral embedding can be integrated into current framework. 
%Also, the explanation of numerical capability learned by large-scale GPT model is still undiscovered, such exploration could be valuable and offer more insights to further research.

%In the future, we plan to extend our methods to a larger size model, as GPT-3 demonstrated its strong performance on numerical computation, enlarging the model size could be a promising direction to improve the numerical capability. Also, the explanation of numerical capability by large-scale GPT model is still undiscovered, such exploration could be valuable and offer more insights to further research.
\bibliography{custom}

\end{document}